# From RAGs to riches: Utilizing large language models to write documents for clinical trials

**Nigel Markey[1], Ilyass El-Mansouri[2], Gaetan Rensonnet[2], Casper van Langen[1] and Christoph Meier[1]**

**Abstract**

**Background/Aims:** Clinical trials require numerous documents to be written: Protocols, consent forms, clinical study reports, and many others. Large language models offer the potential to rapidly generate first-draft versions of these documents; however, there are concerns about the quality of their output. Here, we report an evaluation of how good large language models are at generating sections of one such document, clinical trial protocols.

**Methods:** Using an off-the-shelf large language model, we generated protocol sections for a broad range of diseases and clinical trial phases. Each of these document sections we assessed across four dimensions: *Clinical thinking and logic; Transparency and references; Medical and clinical terminology*; and *Content relevance and suitability.* To improve performance, we used the retrieval-augmented generation method to enhance the large language model with accurate up-to-date information, including regulatory guidance documents and data from ClinicalTrials.gov. Using this retrieval-augmented generation large language model, we regenerated the same protocol sections and assessed them across the same four dimensions.

**Results:** We find that the off-the-shelf large language model delivers reasonable results, especially when assessing *content relevance* and the *correct use of medical and clinical terminology*, with scores of over 80%. However, the off-the-shelf large language model shows limited performance in *clinical thinking and logic* and *transparency and references*, with assessment scores of ≈40% or less. The use of retrieval-augmented generation substantially improves the writing quality of the large language model, with *clinical thinking and logic* and *transparency and references* scores increasing to ≈80%. The retrieval-augmented generation method thus greatly improves the practical usability of large language models for clinical trial-related writing.

**Discussion:** Our results suggest that hybrid large language model architectures, such as the retrieval-augmented generation method we utilized, offer strong potential for clinical trial-related writing, including a wide variety of documents. This is potentially transformative, since it addresses several major bottlenecks of drug development.

## Background and aims

During clinical trials, large volumes of documents need to be written, including protocols, amendments, patient informed consent forms, clinical study reports, and many others. These documents are critically important for the planning and execution of trials and are often required by regulation; therefore, high-quality writing is essential. Specifically, clinical trial documents must be scientifically and clinically precise and accurate, with correct use of terminology, and must contain appropriate references to the literature, regulatory guidelines, and other documents. Due to these stringent requirements, sponsors of clinical trials spend considerable time and resources on trial-related writing. For example, most large pharmaceutical companies each

[1]Boston Consulting Group, London, UK
[2]Boston Consulting Group, Paris, France

**Corresponding author:**
Christoph Meier, Boston Consulting Group, 80 Charlotte Street, London W1T 4DF, UK.
Email: meier.chris@bcg.com

employ tens to hundreds of medical writers and reviewers.[1] Even with these resources, it often takes organizations a long time to write, review, and finalize clinical trial documents. As an illustration, a clinical trial protocol typically has 50–150 or more pages and can take 3–6 months or longer to prepare.[2] A substantial proportion of this time is due to the writing and reviewing process, to ensure that the document achieves the high quality expected. As a result, writing is one of the major rate-limiting steps in the development process. With pharmaceutical companies under pressure to accelerate trials[3,4] and to submit regulatory documents faster, there is strong interest across the industry in using new technologies and approaches to speed up trial-related writing.

In the past few years, large language models (LLMs), a new class of generative artificial intelligence algorithms, have advanced to a point where they can produce near-human-quality writing.[5] Since the arrival of ChatGPT,[6] the first widely used tool built on LLMs, there has been interest in using these algorithms in the context of clinical trials. Examples include enhancing patient-trial matching,[7] clinical trial planning,[8] assisting in medical writing tasks,[9] and others. While it is early days for these efforts, we see signs of great potential but also challenges, such as accuracy and potential biases in the LLMs, as well as concerns about robustness and reproducibility.[10,11] To help address some of these questions, we describe a framework for document quality evaluation, and we report an analysis of LLMs in the context of clinical trial-related writing.

## Methods

Our assessment focused on GPT-4, one of the leading LLMs available today[12] and utilizing it to generate key sections of clinical trial protocols. The LLM output was subsequently assessed in terms of writing quality. Specifically we analyzed four dimensions: *Clinical thinking and logic*, which measures how closely recommendations from regulatory guidance documents were included in the generated section; *Transparency and references*, which verifies the presence and relevance of cited scientific sources in the generated text; *Medical and clinical terminology*, which assesses the use of appropriate jargon and scales of measurements; *Content relevance and suitability*, which measures, among others, whether the generated protocol section was specific to the disease and trial phase. An overview of the methodology is given in Figure 1, and a full description of both the generation and evaluation process, including the criteria and requirements used, is provided in the Supplementary Information.

Our analysis does not make direct comparisons between LLM-written text and fully human-written text. This is because, from our experience in the field, there is often substantial variability between individual human writers. It is therefore challenging to establish a single, objective "ground truth" to compare against. Our evaluation framework, with its four dimensions described above, addresses this challenge by breaking down the assessment into discrete sub-dimensions which can be assessed objectively.

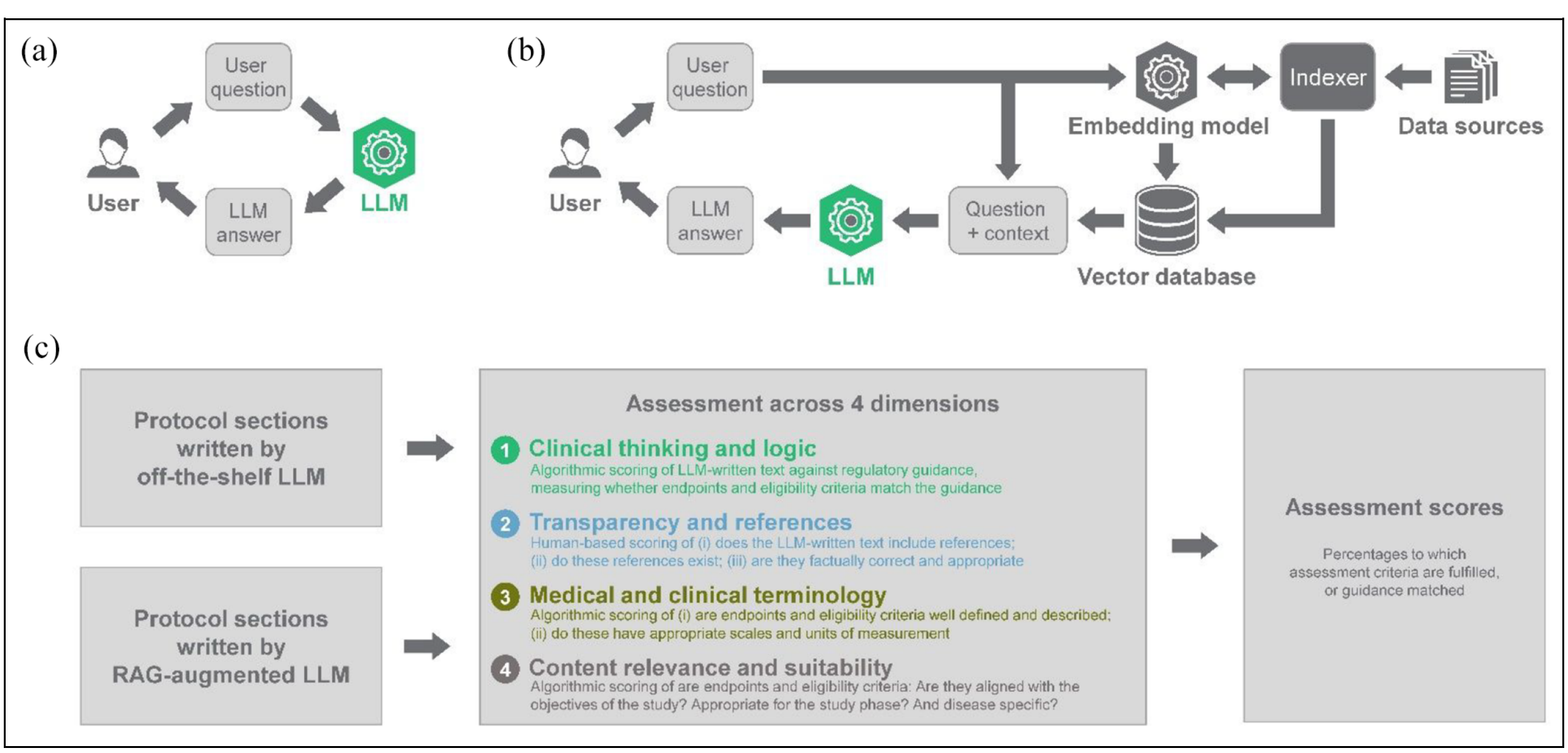

**Figure 1.** Overview of methodology and approach used in this analysis. (a) Typical use of off-the-shelf LLMs. (b) Retrieval-augmented generation (RAG) methodology for enhancing LLMs. (c) ClinEval methodology for assessing the output of large language models (LLMs). Further details are described in the "Methods" section and in the supplementary information.

Our assessment targeted two key sections of a clinical trial protocol document: the *endpoints* section and the *eligibility criteria* section. Two LLM models were evaluated: off-the-shelf GPT-4 via its commercially available application programming interface,[6,12] used as a baseline; and a retrieval-augmented generation (RAG) GPT-4 as an alternative to the off-the-shelf version (see Figure 1(b)). RAG is a methodology for incorporating knowledge from external databases,[13] and involves providing the LLM with external sources of knowledge to supplement the model's internal representation of information.[14] The RAG-augmented LLM was configured as follows: based on a user input query below, an LLM-powered decision agent automatically decided which tools to use to fetch relevant context and feed it to an LLM for final summarization and document generation. As described in more detail in the Supplementary Information, the following tools were utilized as part of the RAG-augmentation: vector store databases to access and analyze regulatory guidance documents; the clinicaltrials.gov AACT (Aggregate Analysis of Clinical Trials) database; SemanticsScholar connector to scrape scientific literature.

Both the off-the-shelf GPT-4 and RAG-augmented GPT-4 were prompted with a natural-language user query of the form "Write the {*section*} section of a Phase {*phase*} clinical trial protocol in {*disease*}. Focus on FDA guidance" where *section*, *phase*, and *disease* were customizable. For each disease and trial phase, and for each model, five endpoints sections and five eligibility criteria sections were generated, with potential differences between versions due to the stochastic nature of the underlying LLM models.

The evaluation process was strictly identical for both the off-the-shelf and RAG-augmented LLMs, and involved a combination of algorithmic and human expert-based scoring. Briefly, the algorithmic assessment consisted in prompting GPT-4, used as an evaluator LLM, with the generated protocol section or an individual section element (i.e. an endpoint or an eligibility criterion) and asking it to provide a binary score for each sub-dimension based on a specific list of requirements (detailed in Supplementary Table 2). For every protocol section that had been generated, metrics for each dimension were then obtained as an average of the scores of the relevant sub-dimensions. Those requirements were developed in consultation with internal and external experts and aim to capture best human practices in clinical protocol writing.

Each of the two models generated a total of 140 document sections, which covered protocols for 14 diseases across different phases of clinical trials (see Supplementary Table 3). The scores are presented as percentages which indicate the mean score achieved by the generated documents across all diseases and phases and section types. Because of the non-deterministic nature of LLMs, we performed five repetitions for each query combination. This approach mitigates the impact of inherent randomness in the model's responses. Statistical tests were performed to assess whether differences in performance metrics between the two models were statistically significant.

## Results

An overview of the result of our assessment is shown in Figure 2. Overall, we find that the off-the shelf LLM delivers reasonable results, specifically good *content relevance and suitability* (assessment score $> 80\%$), and excellent *medical and clinical terminology* ($>99\%$), meaning that the results from the first pass of the LLM are deemed correct and appropriate for the vast majority of the protocol sections written. However, for *clinical thinking and logic*, the off-the-shelf LLM scores poorly (assessment score just over 40%), meaning that recommendations from the off-the-shelf LLM often do not follow the latest regulatory guidance or contain other errors. Since the LLM used for this analysis[6] does not natively source references, assessing *transparency and references* is not possible (therefore, no score for this dimension).

| Assessment dimension | Off-the-shelf LLM | RAG-augmented LLM |
|---|---|---|
| **1 Clinical thinking and logic**<br>Does the LLM-written text reflect regulatory guidance?<br>Do endpoints and eligibility criteria match guidance? | 41.4%* | 79.7%* |
| **2 Transparency and references**<br>Does the LLM-written text include references? Do these exist?<br>Are they factually correct, appropriate? | n/a | 78.6% |
| **3 Medical and clinical terminology**<br>Are endpoints and eligibility criteria defined and described?<br>Do these have scales and units? | >99% | >99% |
| **4 Content relevance and suitability**<br>Are endpoints & eligibility criteria aligned with study objectives?<br>Appropriate for phase, disease specific? | 82.0%* | 79.1%* |

**Figure 2.** Comparison of off-the-shelf LLM and RAG-augmented LLM. Further information is described in the "Methods" section and supplementary information.
*Difference statistically significant ($p < 0.05$).

As an illustrative example when we asked the algorithm to draft a Phase 3 protocol for tuberculosis the off-the-shelf LLM suggested in the eligibility section to exclude patients with human immunodeficiency virus (HIV)/acquired immunodeficiency syndrome, diabetes, liver disease, and kidney disease. This contrasts with regulatory guidance documents which state that "Sponsors should include in trials […], subjects with renal insufficiency, diabetes mellitus, and subjects with hepatic impairment, if feasible. Because of the high incidence of tuberculosis in patients coinfected with HIV, subjects with HIV should be included in trials."[15]

The output of the RAG-augmented LLM (Figure 2) shows high *content relevance and suitability* and *medical and clinical terminology*, comparable to the off-the-shelf LLM. However, the RAG-augmented LLM substantially outperforms the off-the-shelf LLM in terms of *clinical thinking and logic*, where the output of the RAG-augmented LLM scores approximately twice as high as the off-the-shelf LLM. Regarding *transparency and references*, the RAG-augmented LLM (by design) includes references, which we show to be correct and appropriate nearly 80% of the time.

As detailed in Supplementary Table 2, those trends were similar across different protocol sections, with a marked improvement in *clinical thinking and logic*, and similar scores in *medical and clinical terminology* and *content relevance and suitability* (the difference in scores for endpoints was not statistically significant). *Transparency and references* scores of the RAG-augmented LLM were comparable for both protocol sections. Please note that if Bonferroni correction is applied (to account for multiple comparisons), only those related to *clinical thinking and logic* survive. On that dimension, we observed a large difference between off-the shelf LLM and RAG-augmented LLM for the endpoints section (score uplift from ~50% to ~73%) and even more pronounced for the eligibility criteria section (score uplift from ~33% to ~86%).

## Discussion

Across both endpoints and eligibility criteria sections, we find that the off-the-shelf LLM produces seemingly well-written content, as reflected by high scores in *content relevance and suitability* and medical and clinical terminology. Closer investigation of the protocol's specifics, however, reveals important deviations from regulatory guidance (such as the above shown example), captured by low *clinical thinking and logic* scores. Given the critical importance of following regulation, our findings could present a challenge to the use of some off-the-shelf LLMs in the context of clinical trials and may limit their adoption in trial-related document writing. Another major limitation of some off-the-shelf LLMs is their lack of proper referencing.

To address these challenges, we explored alternative approaches of using LLMs, specifically RAG which has emerged as a promising methodology for incorporating knowledge from external databases.[13,14] RAG involves providing the LLM with external sources of knowledge, to supplement the model's internal representation of information.[14] As a result of the RAG methodology, the LLM is primarily used not for its memorized knowledge, but instead for its ability to read, synthesize, and evaluate information provided to it.

In our assessment, the use of RAG augmentation produced high scores for both *clinical thinking and logic* and for *transparency and references*, demonstrating the strength of LLMs and their ability to go beyond writing tasks and reason on novel information provided via RAG. The improvement obtained by the RAG-augmented LLM for *clinical thinking and logic* was particularly high for eligibility criteria. This demonstrates the remarkable ability of RAG-augmented LLMs to exploit vector store databases to find relevant pieces of information within large and complex source documents.

While the results shown in Figure 2 are intriguing, it is important to acknowledge a number of limitations of this work: First, the evaluation framework we report is a mixture of quantitative scores (*clinical thinking and logic*; and *transparency and references*) and qualitative scores (medical and clinical terminology; and content relevance and suitability). The framework thus addresses—at least in part—one of the challenges of assessing clinical trial-related writing, namely, the lack of an objective "ground truth." Going forward, it may be helpful to evolve our framework and make it fully quantitative. Second, the evaluation framework is, by design, relatively general. In our analysis, we cover specific diseases and trial phases but did not tailor it to particular treatments. In practice, this means that the RAG-augmented LLM approach can generate high-quality first draft versions of documents, which would then require further refinement to align with the treatment or therapy being studied. Third, the assessment we report evaluates document sections independently of one another. As a result, it does not capture interrelationships across the full document. This could be important in complex diseases where the endpoints should depend on the target group, defined by eligibility criteria. In such situations, LLM-based writing approaches would be most valuable if they were able to jointly generate and evaluate endpoints and eligibility criteria. One recent approach for this could be to use agent self-reflection,[16] iterating on all protocol sections. Fourth, our assessment looks at only one LLM, namely, GPT-4. This was motivated by the model's popularity and widespread use today. Going forward, additional LLMs (such as Claude 3 Opus[17]) will need to be studied for comparison. Given the recent progress in the field, we expect newer LLMs to have improved performance characteristics.

In summary, our results suggest that hybrid LLM architectures such as agent-based RAG methodology we used offer strong potential for GenAl-powered clinical trial-related writing, covering potentially a wide variety of documents. This is exciting, since it addresses several major bottlenecks of drug development. Indeed, when we applied the RAG-augmented LLM approach in the context of recent clinical trials, we observed dramatic acceleration. For writing tasks, such as protocols or clinical study reports, the time to generate first draft versions of documents is typically reduced from days or weeks (in the case of fully-human writing), to minutes (when using a RAG-augmented LLM). For the end-to-end document creation process, which normally consists of multiple cycles of writing and review, we observe time reductions of 25%–50% or more, depending on which document is being created. The reason why the time reduction of the end-to-end process is somewhat smaller than for writing alone is because review by human experts is always required.

Beyond the writing abilities of LLMs in clinical trials, which our work demonstrates, there are a number of practical considerations which pharmaceutical companies and other trial sponsors will need to address. First and most importantly, there are questions about the ethical use of LLMs and other GenAI tools in the context of clinical trials. If sponsors wish to utilize these tools in the design of clinical trials and the writing of documents, it is imperative that there be responsible human oversight. Clinical trial experts and medical writers will need to be "in the loop" to ensure that trials are designed and executed safely and following all applicable rules and guidelines. Second, there are regulatory questions. The US FDA and other agencies have outlined their plans to regulate artificial intelligence in medical products, including building relevant infrastructure and technical expertise.[18] As the regulatory framework evolves, sponsors of clinical trials will likely need to adapt their LLM and other tools in clinical trial writing and other processes. Third, there are talent and capability considerations. Hiring and retaining suitable expertise, ideally combining clinical trial knowledge and GenAI technical experience, is critical for these efforts to succeed. A recent white paper, jointly authored by major pharmaceutical companies, highlights talent as a major challenge for the industry.[19] Fourth, there are questions regarding technical readiness. In recent years, many pharmaceutical companies and other trial sponsors have made major investments in their data analytics platforms and in data partnerships.[19] However, from the work reported here, we learned that configuring these technologies, and ingesting, integrating, and analyzing the required data sources is often challenging.

Despite these challenges, our experience of starting to deploy LLMs in a number of real-life settings suggests strong potential of accelerating and improving clinical trial-related writing. These benefits typically require a strong collaboration between medical writers, clinical researchers, data scientists, and data engineers. Organizations that achieve this cross-functional collaboration are already beginning to reap significant acceleration gains. Going forward, we expect these benefits to increase even further. Over time, we therefore expect that sponsors of clinical trials will adopt the LLM technology in their clinical and other writing tasks.

## Acknowledgements

The authors thank their colleagues Dr Jennifer Griffin and Dr Souparno Bhattacharya for their assistance in conducting this assessment.

## Author contributions

N.M.: Methodology, Data processing, LLM and other analysis, and Manuscript writing, review, and editing.
C.v.L.: Methodology, Data processing, LLM and other analysis, and Manuscript writing, review, and editing.
G.R.: Methodology, Data processing, LLM and other analysis, and Manuscript review and editing.
I.E.-M.: Methodology, Data processing, and LLM and other analysis.
C.M.: Methodology, LLM and other analysis, and Manuscript writing, review, and editing.
All authors contributed to the article and approved the submitted version.

## Declaration of conflicting interests

The author(s) declared the following potential conflicts of interest with respect to the research, authorship, and/or publication of this article: The authors of this article are employees of The Boston Consulting Group (BCG), a management consultancy that works with the world's leading companies.

## Funding

The author(s) disclosed receipt of the following financial support for the research, authorship, and/or publication of this article: The research for this article was funded by BCG s Health Care practice and by BCG X, the firm's in-house data science unit.

## ORCID iD

Christoph Meier https://orcid.org/0009-0005-6323-8504

## Supplemental material

Supplemental material for this article is available online.

## References

1. Sharma S. How to become a competent medical writer? *Perspect Clin Res* 2010; 1(1): 33–37.
2. Spence O, Hong K, Onwuchekwa Uba R, et al. Availability of study protocols for randomized trials published in high-impact medical journals: a cross-sectional analysis. *Clin Trials* 2020; 17(1): 99–105.
3. Landray MJ, Haynes R and Reith C. Accelerating clinical trials: time to turn words into action. *Lancet* 2023; 402: 165–168.
4. Markey N, Howitt B, El-Mansouri I, et al. Clinical trials are becoming more complex: a machine learning analysis of data from over 16,000 trials. *Sci Rep* 2024; 14: 3514.
5. Biever C. ChatGPT broke the turing test—the race is on for new ways to assess AI. *Nature* 2023; 619(7971): 686–689.
6. OpenAI. ChatGPT. Software tool based on a large language model, 2023, https://openai.com/index/chatgpt/
7. Jin Q, Wang Z, Floudas CS, et al. Matching patients to clinical trials with large language models. *arXiv* 2023; 2307.15051, https://arxiv.org/pdf/2307.15051
8. Wang Z, Xiao C and Sun J. AutoTrial: prompting language models for clinical trial design. *arXiv* 2023; 2305.11366, https://arxiv.org/pdf/2305.11366
9. Patel SB and Lam K. ChatGPT: the future of discharge summaries. *Lancet Digit Health* 2023; 5(3): e107–e108.
10. Huynh LM, Bonebrake BT, Schultis K, et al. New artificial intelligence ChatGPT performs poorly on the 2022 self-assessment study program for urology. *Urol Pract* 2023; 10(4): 409–415.
11. Ghim JL and Ahn S. Transforming clinical trials: the emerging roles of large language models. *Transl Clin Pharmacol* 2023; 31(3): 131–138.
12. OpenAI. GPT-4 technical report. *arXiv* 2023; abs/2303.08774, https://arxiv.org/pdf/2303.08774
13. Gao Y, Xiong Y, Gao X, et al. Retrieval-augmented generation for large language models: a survey. *arXiv* 2023; 2312.10997, https://arxiv.org/pdf/2312.10997
14. Lewis P, Perez E, Piktus A, et al. Retrieval-augmented generation for knowledge-intensive NLP tasks. *Neural Inf Process Syst* 2020; 793: 9459–9474.
15. Department of Health and Human Services, Food and Drug Administration and Center for Drug Evaluation and Research (CDER). Pulmonary tuberculosis: developing drugs for treatment, 2022, https://www.fda.gov/media/87194/download
16. Renze M and Guven E. Self-reflection in LLM agents: effects on problem-solving performance. *arXiv* 2024; 2405.06682, https://arxiv.org/abs/2405.06682
17. Anthropic. Claude 3 Opus [Large language model], 2023. https://www.anthropic.com/
18. US FDA. Using artificial intelligence & machine learning in the development of drug and biological products. *Discussion paper*, 2023, https://www.fda.gov/media/167973/download
19. Khan NS, Senderovitz T, Weatherall J, et al. Data science in pharmaceutical R&D: the DISRUPT-DS industry roundtable. *Nat Rev Drug Discov* 2024; 23(9): 645–646.